\documentclass[10pt,letterpaper]{article}

\usepackage{cogsci}

\usepackage{times}
\usepackage{helvet}
\usepackage{courier}
\usepackage{amssymb}
\usepackage{amsfonts}
\usepackage{amsmath,bm}
\usepackage{enumerate}
\usepackage{mathtools}  
\usepackage{graphicx}
\usepackage{tikz}
\usetikzlibrary{shapes,arrows,automata,fit}
\usepackage{latexsym}
\usepackage[ruled,linesnumbered]{algorithm2e}
\usepackage{url}
\usepackage[caption=false,font=footnotesize]{subfig}
\usepackage{cite}
\usepackage{thmtools}  

\usepackage{enumitem} 
\setitemize[1]{itemsep=0pt,partopsep=0pt,parsep=\parskip,topsep=2pt}  
\setenumerate[1]{itemsep=0pt,partopsep=0pt,parsep=\parskip,topsep=2pt,label={\rm [P\arabic*]},leftmargin=8mm}
\usepackage{bibspacing}  
\makeatletter
\renewcommand\normalsize{%
   \@setfontsize\normalsize\@xpt\@xiipt
   \abovedisplayskip 1\p@ \@plus2\p@ \@minus5\p@
   \abovedisplayshortskip \z@ \@plus3\p@
   \belowdisplayshortskip 5\p@ \@plus3\p@ \@minus3\p@
   \belowdisplayskip \abovedisplayskip
   \let\@listi\@listI}
\makeatother

\title{Attacker and Defender Counting Approach for Abstract Argumentation\thanks{This work was supported by the Funds NSFC 61171121.}}

\author{{\large \bf Fuan Pu, Jian Luo, Yulai Zhang, and Guiming Luo} \\
  School of Software, Tsinghua University, Beijing, China \\
  \{pfa12,j-luo10,zhangyl08\}@mails.tsinghua.edu.cn, gluo@mail.tsinghua.edu.cn
  }

\begin{document}

\newtheorem{theorem}{Theorem}{\bfseries}{\rmfamily}
\newtheorem{definition}{Definition}{\bfseries}{\rmfamily}
\newtheorem{axiom}{Axiom}{\bfseries}{\rmfamily}
\newtheorem{example}{Example}{\itshape}{\rmfamily}
\renewcommand\thmcontinues[1]{Continued}
\newtheorem{exercise}{Exercise}{\itshape\bfseries}{\rmfamily}
\newtheorem{lemma}{Lemma}{\bfseries}{\rmfamily}
\newtheorem{note}{Note}{\bfseries}{\rmfamily}
\newtheorem{problem}{Problem}{\bfseries}{\rmfamily}
\newtheorem{property}{Property}{\bfseries}{\rmfamily}
\newtheorem{proposition}{Proposition}{\bfseries}{\rmfamily}
\newtheorem{corollary}{Corollary}{\bfseries}{\rmfamily}
\newtheorem{conjecture}{Conjecture}{\itshape}{\rmfamily}
\newtheorem{question}{Question}{\bfseries}{\rmfamily}
\newtheorem{remark}{Remark}{\bfseries}{\rmfamily}
\newtheorem{notation}{Notation}{\itshape}{\rmfamily}

\maketitle

\begin{abstract}
In Dung's abstract argumentation, arguments are either acceptable or unacceptable, given a chosen notion of acceptability. This gives a coarse way to compare arguments. In this paper, we propose a counting approach for a more fine-gained assessment to arguments by counting the number of their respective attackers and defenders based on argument graph and argument game. An argument is more acceptable if the proponent puts forward more number of defenders for it and the opponent puts forward less number of attackers against it. We show that our counting model has two well-behaved properties: normalization and convergence. Then, we define a counting semantics based on this model, and investigate some general properties of the semantics.

\end{abstract}

\vspace{-2mm}
\section{Introduction}
Argumentation is an important cognitive process for dealing with conflicting knowledge based on the construction and evaluation of interacting arguments \cite{ref-rahwan2009argumentation}. It has been applied in various domains and applications such as decision making and e-participation. The most popularly used framework to talk about general issues of argumentation is that of Dung's abstract argumentation \cite{ref-Dung1995AAF}, which consists of a set of arguments and a binary relation that represents the conflicting arguments. A number of argumentation semantics for abstract argumentation frameworks have been proposed that highlight different aspects of argumentation \cite{ref-semantic:KER}, such as admissible sets, preferred extension, and grounded extension. However, these semantics provide a rather rough way to evaluate arguments and may result in some undesired results \cite{ref-bench2007ArgAI}. A common case is that a semantics may give an empty answer. Conversely, several answers may be provided, with nothing to distinguish between them.

In order to overcome these difficulties, there is a trend towards considering and exploring the possibility of discriminating between arguments by employing a larger number of categories or continuous numerical scales \cite{ref-cayrol2005graduality,ref-matt2008game,ref-leite2011SocialAF,ref-gabbay2012equational,modgil2013added}. One of the main advantages of these works is that it allows for a more fine-grained assessment on arguments than is provided by the traditional extensions-based approaches. We aim at following these works by evaluating the strength of arguments on a scale of numerical values from $0$ to $1$ so as to finely compare and rank arguments from the most acceptable to the weakest one(s). This fits well with recent interest in quantitative measures for the ranking analysis of argumentation \cite{ref-amgoud2013ranking,ref-pfa2014caterank}.

In this paper, our fundamental idea used to formalise argument strength is essentially the same as those found in abstract argumentation theory: argument $x$ is more acceptable than argument $y$ iff $x$ has a better defence (for it) and a lower attack (against it). In order to assess the strength of arguments in an argumentation framework, we will consider their evaluation procedures as dialogue games \cite{Simari2009argame}, where two fictitious agents---one \textsf{PRO} (the proponent) and the other \textsf{OPP} (the opponent)---take part in. A dialogue game begin with \textsf{PRO} putting forward an initial argument, and then \textsf{PRO} and \textsf{OPP} take turns in a sequence of moves called a \emph{dispute}, in which each agent makes an argument that attacks its counterpart's last move. In general, the counterpart can try a different line of attack and create a new dispute. This leads to a dispute tree structure that represents the dialogue game. Nodes in a dispute tree are labelled by arguments and are assigned the status of \emph{defender node} and \emph{attacker node} of the root argument, depending upon the argument at that node is made by the proponent or by the opponent, or depending upon whether the walk length between the current node and the root node is even or odd. We claim that an argument is more acceptable if \textsf{PRO} puts forward more number of defender nodes for it and \textsf{OPP} puts forward less number of attacker nodes against it. We will thus introduce a graded approach to assess the strength of each argument based on its dispute tree by counting its defender nodes and attacker nodes.

The rest of this paper is organized as follows. In Section 2 we briefly recall some background on Dung's abstract argumentation and argument game. We present the attacker and defender counting semantics in Section 3. Some properties of the semantics are investigated in Section 4. Section 5 discusses related work and concludes.

\section{Preliminaries}
\subsection{Abstract argumentation framework}
We consider the basic concepts and insights of Dung's abstract argumentation framework, in which both arguments and attacks are assumed to be abstract entities \cite{ref-Dung1995AAF}.
\begin{definition}[Argumentation framework]
\label{Def_Argumentation_Framework}
An \emph{argumentation framework} (or AF, in short) is a pair $\textit{AF}=\left< \mathcal{X}, \mathcal{R}\right>$ where $\mathcal{X}$ is a set of arguments and $\mathcal{R} \subseteq \mathcal{X} \times \mathcal{X}$ is a binary relation called \emph{attack relation}. For two arguments $x,y\in\mathcal{X}$, $(x,y)\in\mathcal{R}$ or $x\mathcal{R}y$ means that $x$ attacks $y$.
\end{definition}

We denote by $\mathcal{R}^-(x)$ (respectively, $\mathcal{R}^+(x)$) the subset of $\mathcal{X}$ containing those arguments that attack (respectively, are attacked by) the argument $x\in\mathcal{X}$, extending this notation in the natural way to sets of arguments, so that for $S\subseteq \mathcal{X}$, $\mathcal{R}^-(S) \triangleq \{x\in\mathcal{X}: \exists y \in S \mbox{~such that~} x\mathcal{R}y\}$ and $\mathcal{R}^+(S) \triangleq \{x\in\mathcal{X}: \exists y \in S \mbox{~such that~} y\mathcal{R}x\}$.

To define the solutions of an AF, we mean selecting a set of arguments that satisfy some acceptable criteria. Dung presents several of these properties, called extensions or semantics, which produce zero, one, or several sets of accepted arguments. These semantics are based on two important concepts: conflict-freeness and defence.
\begin{definition}[Conflict-free, Defense] \label{Def_CF&D}
Let $\textit{AF}=\left< \mathcal{X}, \mathcal{R}\right>$ be an argumentation framework, let $S\subseteq\mathcal{X}$ and $x\in\mathcal{X}$.
\begin{itemize}
  \item $S$ is \emph{conflict-free} iff $\nexists x,y\in S$ such that $x\mathcal{R}y$. 
  \item $S$ \emph{defends} argument $x$ iff $\forall y\in\mathcal{X}$ if $y\mathcal{R}x$ then $\exists z\in S$ such that $z\mathcal{R}y$. It is also said that argument $x$ is \emph{acceptable} with respect to $S$. 
\end{itemize}
\end{definition}
Using the notions of conflict-freeness and defence, we can define a number of argumentation semantics, each embodying a particular rationality criterion.
\begin{definition}[Acceptability semantics]
Let $S\subseteq\mathcal{X}$ be conflict-free, and let $\mathfrak{F}: 2^\mathcal{X}\mapsto 2^\mathcal{X}$ be a function such that $\mathfrak{F}(S)\triangleq \{x\in\mathcal{X}: S ~\mbox{defends}~ x\}$.
\begin{itemize}
  \item $S$ is \emph{admissible} iff $S\subseteq\mathfrak{F}(S)$.
  \item $S$ is a \emph{preferred extension} iff it is a maximal (w.r.t $\subseteq$) admissible set.
  \item $S$ is a \emph{stable extension} iff it attacks all arguments in $\mathcal{X} \backslash S$. 
  \item $S$ is a \emph{complete extension} iff $S=\mathfrak{F}(S)$.
  \item $S$ is a \emph{grounded extension} iff it is the minimal (w.r.t $\subseteq$) complete extension.
\end{itemize}
\end{definition}

\begin{example} \label{Exp_SimpleAF}
  Consider an $\textit{AF}=\left< \mathcal{X}, \mathcal{R}\right>$, described in Figure~\ref{Fig_AAF}, in which $\mathcal{X}=\{x_1, x_2, x_3, x_4\}$ and $\mathcal{R}=\{(x_2,x_1),(x_3,x_2),(x_2,x_3),(x_3,x_3), (x_4,x_2)\}$. For this example, $\textit{AF}$ has three admissible sets:  $\emptyset$, $\{x_4\}$ and $\{x_1, x_4\}$. $\{x_1,x_4\}$ is the only preferred extension of $\textit{AF}$, and it is also complete and grounded. $\textit{AF}$ has no stable extension.
\end{example}

\subsection{Argument graph and argument game}
An argumentation framework can be represented as a digraph, called \emph{argument graph}, in which vertices are arguments and directed arcs characterise attack relations between arguments.
\begin{definition}
Let $\mathbb{G}$ be the argument graph associated to the argumentation framework $\left< \mathcal{X}, \mathcal{R}\right>$:
\begin{itemize}
  \item A \emph{walk} from $x$ to $y$ is a sequence of arguments $\mathcal{S}=\left<x'_0,x'_1,\cdots,x'_m\right>$ such that $x'_0 = x$, $x'_m=y$ and $x'_{t-1}\mathcal{R}x'_t$ for all $t \in \{1,2,\cdots,m\}$. The \emph{length} of this walk, denoted by $\ell_\mathcal{S}$, is the number of edges used in the walk. We denote the set of all walks from $x$ to $y$ of length $\ell$ by $\mathcal{S}(x,y,\ell)$.
  \item A \emph{cycle} is a walk $\mathcal{S}=\left<x'_0,x'_1,\cdots,x'_{m-1},x'_0\right>$. A cycle is an \emph{elementary cycle} iff for any $i,j\in\{1,2,\cdots,m-1\}$ such that if $i\neq j$ then $x'_i\neq x'_j$.
  \item $\mathbb{G}$ is \emph{acyclic} iff there are no cycles in $\mathbb{G}$.
\end{itemize}
\end{definition}

In this paper, we assess the strengths of arguments based on abstract argument games without regard to the specific internal structure of the arguments \cite{Simari2009argame}. These games typically assume the presence of two fictitious agents, \textsf{PRO} (for ``proponent'') and \textsf{OPP} (for ``opponent''). Each game start with \textsf{PRO} asserting an initial argument to be tested. \textsf{OPP} and \textsf{PRO} then take turns in moving arguments that successively attack each other's last move. A sequence of moves in which each agent moves against its counterpart's last move is referred to as a \emph{dispute}. Generally, however, an agent can backtrack to a counterpart's previous move and initiate a new dispute. Thus, the data structure of an argument game can be represented by an argument graph's induced \emph{dispute tree}, in which each branch from root to leaf is a dispute:
\begin{definition}[Dispute tree]
  Let $\textit{AF}=\left< \mathcal{X}, \mathcal{R}\right>$ be an argumentation framework, and let $x\in\mathcal{X}$. The \emph{dispute tree} induced by $x$ in $\textit{AF}$ is a tree $\mathbb{T}$ of arguments, such that the root of $\mathbb{T}$ is $x$, and for any $y,z\in\mathcal{X}$, $y$ is a child of $z$ in $\mathbb{T}$ iff $y\mathcal{R}z$.
\end{definition}

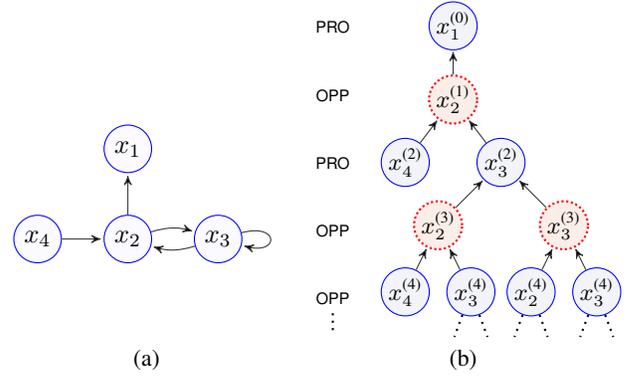
\begin{figure}[tb]
\centering
\subfloat[]{
\begin{tikzpicture}[->,>=stealth',shorten >=1pt,auto,node distance=1.2cm, thin]
\tikzstyle{vNorm}=[draw=blue,fill=blue!2,circle,text width=5mm,inner sep=1pt,minimum height=6pt, align=center]
\tikzstyle{vBlank}=[text width=5mm, font={\tiny\sf}, align=center] 
\tikzstyle{every edge}=[draw=black!80, font=\scriptsize]

\node[vNorm](x1){$x_1$};
\node[vNorm, below of=x1](x2){$x_2$};
\node[vNorm, right of=x2](x3){$x_3$};
\node[vNorm, left of=x2](x4){$x_4$};

\path (x2) edge               (x1)
      (x2) edge [bend left=17](x3)
      (x3) edge [bend left=17](x2)
      (x3) edge [loop right]  (x3)
      (x4) edge               (x2);

\node[vBlank, below of=x2](B1){};

\end{tikzpicture}
\label{Fig_AAF}}
\hfil
\subfloat[]{
\begin{tikzpicture}[auto,node distance=0.9cm]
\definecolor{Attacker}{rgb}{0.85,0.33,0.1}
\definecolor{Defender}{rgb}{0.08,0.17,0.55}
\tikzstyle{vBlue}=[draw=blue,fill=Defender!5, thin, circle,inner sep=0.2pt, minimum height=6pt, text width=5mm, font=\footnotesize, align=center]
\tikzstyle{vRed}=[draw=red,densely dotted,thick, fill=Attacker!10,circle,inner sep=0.2pt, minimum height=6pt, text width=5mm, font=\small, align=center]
\tikzstyle{vTxt}=[text width=5mm, font={\tiny\sf}, align=center]
\tikzstyle{every edge}=[draw=black!80, ->,>=stealth',shorten >=0.5pt, font=\footnotesize]

\node[vTxt](P1){PRO};
\node[vTxt, below of=P1](O1){OPP};
\node[vTxt, below of=O1](P2){PRO};
\node[vTxt, below of=P2](O2){OPP};
\node[vTxt, below of=O2](P3){OPP};
\draw[dotted, color=black, thick] (node cs:name=P3) -- +(0,-4.5mm);

\node[vBlue, right of=P1, xshift=7mm, yshift=0.2mm](x11){$x_{\scriptstyle 1}^{\mbox{\tiny (0)}}$};
\node[vRed, below of=x11, yshift=-0.8mm](x21){$x_{\scriptstyle 2}^{\mbox{\tiny (1)}}$};
\node[vBlue, below left of=x21, yshift=-2mm](x31){$x_{\scriptstyle 4}^{\mbox{\tiny (2)}}$};
\node[vBlue, below right of=x21, yshift=-2mm](x32){$x_{\scriptstyle 3}^{\mbox{\tiny (2)}}$};
\node[vRed, below left of=x32, yshift=-2mm, xshift=-2mm](x41){$x_{\scriptstyle 2}^{\mbox{\tiny (3)}}$};
\node[vRed, below right of=x32, yshift=-2mm, xshift=2mm](x42){$x_{\scriptstyle 3}^{\mbox{\tiny (3)}}$};
\node[vBlue, below left of=x41, yshift=-2.5mm, xshift=2mm](x51){$x_{\scriptstyle 4}^{\mbox{\tiny (4)}}$};
\node[vBlue, below right of=x41, yshift=-2.5mm, xshift=-2mm](x52){$x_{\scriptstyle 3}^{\mbox{\tiny (4)}}$};
\node[vBlue, below left of=x42, yshift=-2.5mm, xshift=2mm](x53){$x_{\scriptstyle 2}^{\mbox{\tiny (4)}}$};
\node[vBlue, below right of=x42, yshift=-2.5mm, xshift=-2mm](x54){$x_{\scriptstyle 3}^{\mbox{\tiny (4)}}$};
\path (x21) edge               (x11)
      (x31) edge               (x21)
      (x32) edge               (x21)
      (x41) edge               (x32)
      (x42) edge               (x32)
      (x51) edge               (x41)
      (x52) edge               (x41)
      (x53) edge               (x42)
      (x54) edge               (x42);

\draw[dotted, color=black, thick] (node cs:name=x52) -- +(-2.2mm,-6mm);
\draw[dotted, color=black, thick] (node cs:name=x52) -- +(2.2mm,-6mm);
\draw[dotted, color=black, thick] (node cs:name=x53) -- +(-2.2mm,-6mm);
\draw[dotted, color=black, thick] (node cs:name=x53) -- +(2.2mm,-6mm);
\draw[dotted, color=black, thick] (node cs:name=x54) -- +(-2.2mm,-6mm);
\draw[dotted, color=black, thick] (node cs:name=x54) -- +(2.2mm,-6mm);

\end{tikzpicture}
\label{Fig_DisputeTree}}
\caption{Argumentation framework and dispute tree. (a) shows an argumentation framework, (b) shows the dispute tree induced in $x_1$.}
\label{Fig_AF&DisputeTree}
\end{figure}

Nodes in a dispute tree are labelled by arguments and are assigned the status of \emph{defender node} or \emph{attacker node} of the root argument, depending upon whether the walk length from the current node to the root node is even or odd, or depending upon whether the argument at that node is made by \textsf{PRO} or by \textsf{OPP}. Consider two agents arguing the argumentation framework shown in Figure~\ref{Fig_AAF}, and the dispute tree induced by $x_1$ is shown Figure~\ref{Fig_DisputeTree}. Note that this dispute tree is infinite, since both agents are able to repeat counterarguments due to the presence of cycles in the argument graph. In this dispute tree, the blue solid nodes, put forward by \textsf{PRO}, are defender nodes of $x_1$, whereas, the red dotted nodes, made by \textsf{OPP}, are attacker nodes of $x_1$. Each node is also assigned a superscript, which denotes the length of the move sequence from the current node to the root node. Obviously, if a node has a even-numbered superscript then it is a defender node, otherwise it is an attacker node. Note that the root node is also a defender node of $x_1$ since each argument has a walk with length $0$ to itself.


In this paper, we define the argument within a defender (respectively, attacker) node is a defender (respectively, attacker) of the argument within the root node. An argument $x$ is a defender or an attacker of argument $y$ depending on the length of the walk between them. Now, let us define attacker and defender based on argument graph:
\begin{definition}[Attacker and Defender]
Let $\left< \mathcal{X}, \mathcal{R}\right>$ be an AF, and $\mathbb{G}$ be its argument graph. Let arguments $x,y\in\mathcal{X}$.
\begin{itemize}
  \item $x$ is an \emph{attacker} of $y$ if there exists a walk $\mathcal{S}$ from $x$ to $y$ such that $\ell_\mathcal{S}=2t+1$ with $t=0,1,2,\cdots$. Then, $x$ is said to be a $\ell_\mathcal{S}$-length attacker of $y$.
  \item $x$ is an \emph{defender} of $y$ if there exists a walk $\mathcal{S}$ from $x$ to $y$ such that $\ell_\mathcal{S}=2t$ with $t=0,1,2,\cdots$. Then, we call $x$ is said to be a $\ell_\mathcal{S}$-length defender of $y$.
\end{itemize}
\end{definition}

Note that an defender can also be a attacker (e.g., an defender node and an attacker node are labelled by the same argument). In the same way, two defenders can be the same argument (e.g., two different lengths of defender nodes are labelled by the same argument) and the same thing may occur for the attackers. In \cite{ref-cayrol2005graduality}, the authors distinguish attackers (respectively, defenders) by direct and indirect. In this paper, instead, we distinguish them by different walk and walk length. Accordingly, if there exists $m$ number of $\ell$-length walks from argument $x$ to argument $y$, i.e. $|\mathcal{S}(x,y,\ell)|=m$, then we consider that $x$ is the $m$ number of different $\ell$-length attackers or defenders of $y$.

\begin{example} \label{Exp_Attacker&Defender}
  Consider the argumentation graph depicted in Figure~\ref{Fig_AF&DisputeTree}. It can be easily see that there are two elementary cycles $\left<x_2,x_3,x_2\right>$ and $\left<x_3,x_3\right>$. Since $\mathcal{S}(x_2,x_1,1)=\{\left<x_2,x_1\right>\}$ and $\mathcal{S}(x_2,x_1,3)=\{\left<x_2,x_3,x_2,x_1\right>\}$, thus $x_2$ is a $1$-length and $3$-length attacker of $x_1$ (corresponding to the attacker nodes $x_2^{(1)}$ and $x_2^{(3)}$ in the dispute tree). Note that $x_2$ is also a defender of $x_1$ due to the $4$-length walk $\left<x_2,x_3,x_3,x_2,x_1\right>$ (corresponding to the defender node $x_2^{(4)}$). There exist two walks from $x_3$ to $x_1$ of length $4$, i.e., $\mathcal{S}(x_3,x_1,4)=\{\left<x_3,x_3,x_3,x_2,x_1\right>,\left<x_3,x_2,x_3,x_2,x_1\right>\}$ (corresponding to two defender nodes $x_3^{(4)}$), thus $x_3$ is two different $4$-length defenders of $x_1$.
\end{example}

\section{Attacker and Defender Counting Semantics}
In classical abstract argumentation, arguments are either acceptable or unacceptable, given a chosen notion of acceptability. This gives a rather coarse way to compare arguments. In this paper, we intend to provide a more fine-grained evaluation of arguments based on the graph structure of the argument system. Our basic starting point is that argument $x$ is more acceptable than argument $y$ iff \textsf{PRO} makes more defenders for $x$ and \textsf{OPP} makes less attackers against $x$.

Towards such an idea, our approach is to count the number of all attackers and defenders for each argument. The less the attackers and the more defenders an argument has, the more acceptable the argument. In this approach, the main constraint is that we must be able to identify all attackers and defenders for each argument. This is quite easy in the case of argument graphs without cycles. In this section, we will introduce first a matrix approach to record and track all attackers and defenders of different lengths for every argument regardless of whether the argument graph is acyclic or cyclic. Then, counting models are established to assess the strengths of arguments, and a counting semantics is defined. The properties of this semantics are studied in the next section.

\subsection{Finding attackers and defenders}
In this subsection, we will use a series of matrices to memorise the number of all walks with different lengths between
any two arguments, and will present a matrix product approach to compute these matrices.

Let $\textit{AF}=\left< \mathcal{X}, \mathcal{R}\right>$ be an argumentation framework with $\mathcal{X}=\{x_1,x_2,\cdots,x_n\}$. We use a $n\times n$ matrix $\bm{A}^{(\ell)}=[a^{(\ell)}_{ij}]$ to memorise the number of $\ell$-length walks between any pair of arguments, which is defined as
\begin{equation*}
  a^{(\ell)}_{ij} = |\mathcal{S}(x_j,x_i,\ell)|
\end{equation*}
Intuitively, $\bm{A}^{(0)}=\bm{I}$ where $\bm{I}$ is the identity matrix. Now let us define another $n\times n$ matrix $\bm{A}=[a_{ij}]$, called \emph{attack matrix}, where entry $a_{ij}$ is $1$ iff $x_j\mathcal{R}x_i$; otherwise $0$. Obviously, $\bm{A}$ is the transpose of the adjacency matrix of the attack graph of $\textit{AF}$. Then, it is easy to see that $\bm{A}^{(1)}=\bm{A}$, and further we have the following result:
\begin{lemma} \label{Lmm_Matrix&Walk}
  $\bm{A}^{(\ell)} = \bm{A}^\ell$ for each $\ell=0,1,2,\cdots$.
\end{lemma}

This lemma is common in graph theory. We introduce it here since it provides us a way to represent and compute the number of all walks with different lengths between any two arguments. It implies that by this approach we can find all attackers or defenders of various lengths of an argument. From now on, we will use $\bm{A}^\ell$ to substitute for $\bm{A}^{(\ell)}$ for brevity.
\begin{proposition} \label{Prop_Cycle&Acyclic}
Let $\mathbb{G}$ be the argument graph of $\textit{AF}=\left< \mathcal{X}, \mathcal{R}\right>$ and its attack matrix be $\bm{A}$.
\begin{enumerate}
  \item If there exists a cycle in $\textit{AF}$, then for any positive integer $\ell$ it holds that $\bm{A}^\ell \neq \mathbf{0}$. \footnote{A matrix $\bm{M} \neq \mathbf{0}$ means that there exists some entry in $\bm{M}$ is non-zero. On the contrary, $\bm{M} = \mathbf{0}$ means that all entries in $\bm{M}$ are zeros.}
  \item If $\mathbb{G}$ is acyclic, then there exists a positive integer $r$ such that $\bm{A}^\ell \neq \mathbf{0}$ for any positive integer $\ell \leq r$, and $\bm{A}^{\ell} = \mathbf{0}$ for any integer $\ell \geq r+1$. Moreover, $r$ is the length of the longest walk in $\mathbb{G}$.
\end{enumerate}
\end{proposition}
This proposition reveals that if the attack graph of an argumentation framework is acyclic, the attackers and defenders of each argument is finite; otherwise, an argument in a cyclic graph may have infinite attackers and defenders.

\subsection{The counting models for argumentation framework}
Now, we will concern on establishing the counting model for the evaluation of arguments.

\subsubsection{The simple counting model}
The basic idea behind the simple counting model is: for each argument $x$, for each walk length $\ell$, to count the number of $\ell$-attackers or $\ell$-defenders for $x$. We positively count all defenders and negatively count all attackers. This is easy to understand since an argument is always weakened by its attackers and is ``reinstated'' by its defenders. Therefore, in any case, the greater the number computed, the more acceptable the argument $x$.

Cycles in argument graphs are expensive as the attackers and defenders of an argument may be infinite. Here we firstly consider the approach to count attackers and defenders under a given maximum walk length, denoted by $k$, which will be used in order to capture finite attackers and defenders. Let $\textit{AF}=\left< \mathcal{X}, \mathcal{R}\right>$ with $\mathcal{X}=\{x_1,x_2,\cdots,x_n\}$ and let $\bm{v}$ be the $n$-dimensional column vector over $\mathcal{X}$. Given the maximum walk length $k$, we define the simple counting model as
\begin{equation} \label{Eqn_SimpleCounting}
  \bm{v}^{(k)}=\sum^{k}_{\ell=0}(-1)^\ell\bm{A}^\ell\bm{e}
\end{equation}
where $\bm{e}$ is the column vector consisting of all ones. Note that here $(-1)^\ell$ encodes the consideration of positively and negatively counting since $(-1)^\ell$ is $-1$ for odd $\ell$ and $1$ for even $\ell$. The item $\bm{A}^{\ell}\bm{e}$ means counting the number of all $\ell$-length attackers or defenders of each argument. As $k$ goes to $\infty$, then $\bm{v}^{(k)}$ is the evaluation on arguments.

However, there are two problems with this simple counting model. The first is that for an attack graph with cycles, when $k$ goes to $\infty$, then some arguments may have infinite number of attackers and defenders, which may cause $\bm{v}^{(k)}$ go to $\infty$. Considering the attack graph in Figure~\ref{Fig_AAF}, for example, there are $3.54\times10^{20}$ different walks from $x_3$ to $x_2$ of length $100$. As the representation and processing of the infinite case is difficult, the simple counting model is not conducive to comparison and practical application. For example, if the counting values of two arguments are both infinite, we can not compare them.

The second problem is that the simple counting model does not distinguish different lengths of attackers and defenders. Different lengths of attackers or defenders of an argument may have different effects on the argument. The simple model just simply counting them together and does not consider which is more important and which is less important. In this paper, shorter attackers and defenders are preferred, which can effectively drive the agents to make only relevant moves, and thus we assume that a shorter attacker (respectively, defender) of an argument has more effect than a longer one on the argument \cite{ref-rienstra2013opponent}. More concretely, considering the move sequence $x_3^{(4)} \rightarrow x_3^{(3)} \rightarrow x_3^{(2)} \rightarrow x_2^{(1)} \rightarrow x_1^{(0)}$ in Figure~\ref{Fig_DisputeTree}, where $x_1^{(0)}$, $x_3^{(2)}$ and $x_3^{(4)}$ are three defender nodes of $x_1$, and $x_2^{(1)}$ and $x_3^{(3)}$ are two attacker nodes of $x_1$. Here, we consider $x_1^{(0)}$ has more (defence) effect than $x_3^{(2)}$ on $x_1$ since $x_3^{(2)}$ has a longer walk to $x_1$. Similar viewpoint gives that $x_3^{(2)}$ has more (defence) effect than $x_5^{(4)}$ on $x_1$, and that $x_2^{(1)}$ has more (attack) effect than $x_3^{(3)}$ on $x_1$.

\subsubsection{The improved counting model}
To remedy these two problems, we firstly define a \emph{normalization factor}, which can ensure that the argument strength scale is bounded, and secondly we define a \emph{damping factor} on walk length, which allows a more refined treatment on different length of attacker and defenders. Then, we write the improved counting model as
\begin{equation} \label{Eqn_ImprovedCounting}
  \bm{v}^{(k)}=\sum^{k}_{\ell=0}(-1)^\ell\alpha^\ell\widetilde{\bm{A}}^\ell\bm{e}
\end{equation}
in which $\alpha\in(0,1)$ is the damping factor and $\widetilde{\bm{A}}$ is the normalized attack matrix defined as $\widetilde{\bm{A}}=\bm{A}/{N}$ where the scalar $N$ is the normalization factor. Now, we can see that the damping factor $\alpha$ provides a graded treatment of attackers and defenders of various lengths since the longer the walk length $\ell$, the smaller the $\alpha^\ell$.

To ensure bounded $\bm{v}^{(k)}$, the underlying principle to select the normalization factor $N$ should satisfy the spectral radius of $\widetilde{\bm{A}}$ no more than $1$ \cite[Chapter.~5]{ref-horn2012matrix}. In this paper, we select $N$ as the matrix infinite norm of $\bm{A}$, defined for $\bm{A}$ by
\begin{equation*}
  N  = \|\bm{A}\|_\infty=\max_{1\leq i \leq n}\sum_{j=1}^{n}|a_{ij}|
\end{equation*}
since it provides two well-behaved properties as follows:

\begin{theorem}[Normalization] \label{Thrm_Normalization}
  For any non-negative integer $k$, the improved counting model $\bm{v}^{(k)}$ defined in Equation~\ref{Eqn_ImprovedCounting} is such that $\mathbf{0} \leq \bm{v}^{(k)} \leq \bm{e}$.
\end{theorem}
For any argumentation system, the improved counting model can range the strength value of each argument into the interval $[0,1]$, as it uses a dynamic normalization factor $N$, in other words, the norm of an attack matrix used here represents the ``size'' of its corresponding argumentation framework. With this normalization property, all arguments can be easily compared. We must note that here the strength values of arguments are relative and not the real number of their attackers and defenders, hence, they do not make sense when they are not compared with each other.

Another property is called \emph{convergence}, which states that as $k$ goes to $\infty$, the improve counting model will converge.
\begin{theorem}[Convergence] \label{Thrm_Convergence}
  The sequence $\{\bm{v}^{(k)}\}^\infty_{k=0}$ defined by Equation~\ref{Eqn_ImprovedCounting} necessarily converges.
\end{theorem}
The proof this theorem needs to consider two cases, i.e., attack graph $\mathbb{G}$ contains cycle(s) or not. We can prove that for both cases, the improved counting model always converges to a unique solution.
\vspace{-1mm}
\subsection{The counting semantics for AF}
We now define the attacker and defender counting semantics for an argumentation framework as the limit of $\{\bm{v}^{(k)}\}_{k=0}^\infty$.
\begin{definition}
\label{Def_ArgStrength}
Let $\textit{AF}=\left< \mathcal{X}, \mathcal{R}\right>$ be an argumentation framework with $\mathcal{X}=\{x_1,x_2,\cdots,x_n\}$. The \emph{attacker and defender counting semantics} for such $\textit{AF}$ is, for all arguments $\mathcal{X}$,
\begin{equation*}
  \bm{v} = \lim_{k\rightarrow\infty} \bm{v}^{(k)}
\end{equation*}
The strength value of each argument $x_i$ is denoted as $\bm{v}(x_i)$.
\end{definition}

To obtain this counting semantics, one basic idea is to compute $\bm{v}^{(0)},\bm{v}^{(1)},\cdots$ until either $\bm{v}^{(k)}=\bm{v}^{(k-1)}$ or the approximation is considered adequate. If directly utilizing Equation~\ref{Eqn_ImprovedCounting}, this may incur prohibitively expensive computational cost since for each $k$ we need to recompute all attackers and defenders for every argument. By Equation~\ref{Eqn_ImprovedCounting}, however, we can easily derive the following iteration approach:
\begin{equation} \label{Eqn_Iteration}
  \bm{v}^{(k)} = \bm{e} - \alpha\widetilde{\bm{A}}\bm{v}^{(k-1)}
\end{equation}
Then, the next valuation can be computed by the outputs of the previous iteration. With the initial valuation $\bm{v}^{(0)}=\bm{e}$, we can approximate the unique solution by iteration. This iterative approach is done by using Algorithm~\ref{Alg_IterativeValuation}. On line \ref{Alg_line:3} we substitute $\widehat{\bm{A}}$ for $\alpha\widetilde{\bm{A}}$ to reduce the calculation, and on line \ref{Alg_line:7} the change $\delta$ is computed. In line \ref{Alg_line:until} the iteration terminates when the change $\delta$ is under a given tolerance $\epsilon$. It can be proved that the convergence speed of this iteration algorithm is linear and no more than $\alpha$.

\begin{algorithm}[htb] \label{Alg_IterativeValuation}
\KwIn{$\alpha$: damping factor; $\bm{A}$: attack matrix\;
~~~~~~~~~~~~$\epsilon$: prescribed tolerance\;}
\KwOut{$\bm{v}^{(k)}$: the approximate counting semantics}
$k \longleftarrow 0$;
$\bm{v}^{(0)} \longleftarrow \bm{e}$\;
$\widehat{\bm{A}} \longleftarrow \alpha\bm{A}/\|\bm{A}\|_\infty$\;  \label{Alg_line:3}

\Repeat{$\delta \leqslant \epsilon$}{
$k \longleftarrow k+1$\;
$\bm{v}^{(k)} \longleftarrow \bm{e}-\widehat{\bm{A}}\cdot\bm{v}^{(k-1)}$\;
$\delta = \|\bm{v}^{(k)}-\bm{v}^{(k-1)}\|$\; \label{Alg_line:7}
}\label{Alg_line:until}
\Return $\bm{v}^{(k)}$\;
\caption{An Iteration Approach for Attacker and Defender Counting Semantics}
\end{algorithm}

\begin{example}  \label{Exp_ValuationSeq}
Consider again the argumentation framework in Example~\ref{Exp_SimpleAF}. Let $\alpha=0.98$ and $\epsilon = 10^{-3}$. Then, the valuation sequence of the attacker and defender counting model, calculated by Algorithm~\ref{Alg_IterativeValuation}, is shown in Figure~\ref{Fig_IterValuation}. The valuation sequence reflects how the strength value of each argument changes with various maximum walk length $k$. After finitely many iterations, the valuation sequence gradually tends to be stable and converges to the approximative counting semantics $\bm{v}=[0.89, 0.22, 0.60, 1.00]^T$ within a tolerable range.
\begin{figure}[htb]
\vspace{-3mm}
\centering
  \includegraphics[width=0.48\textwidth]{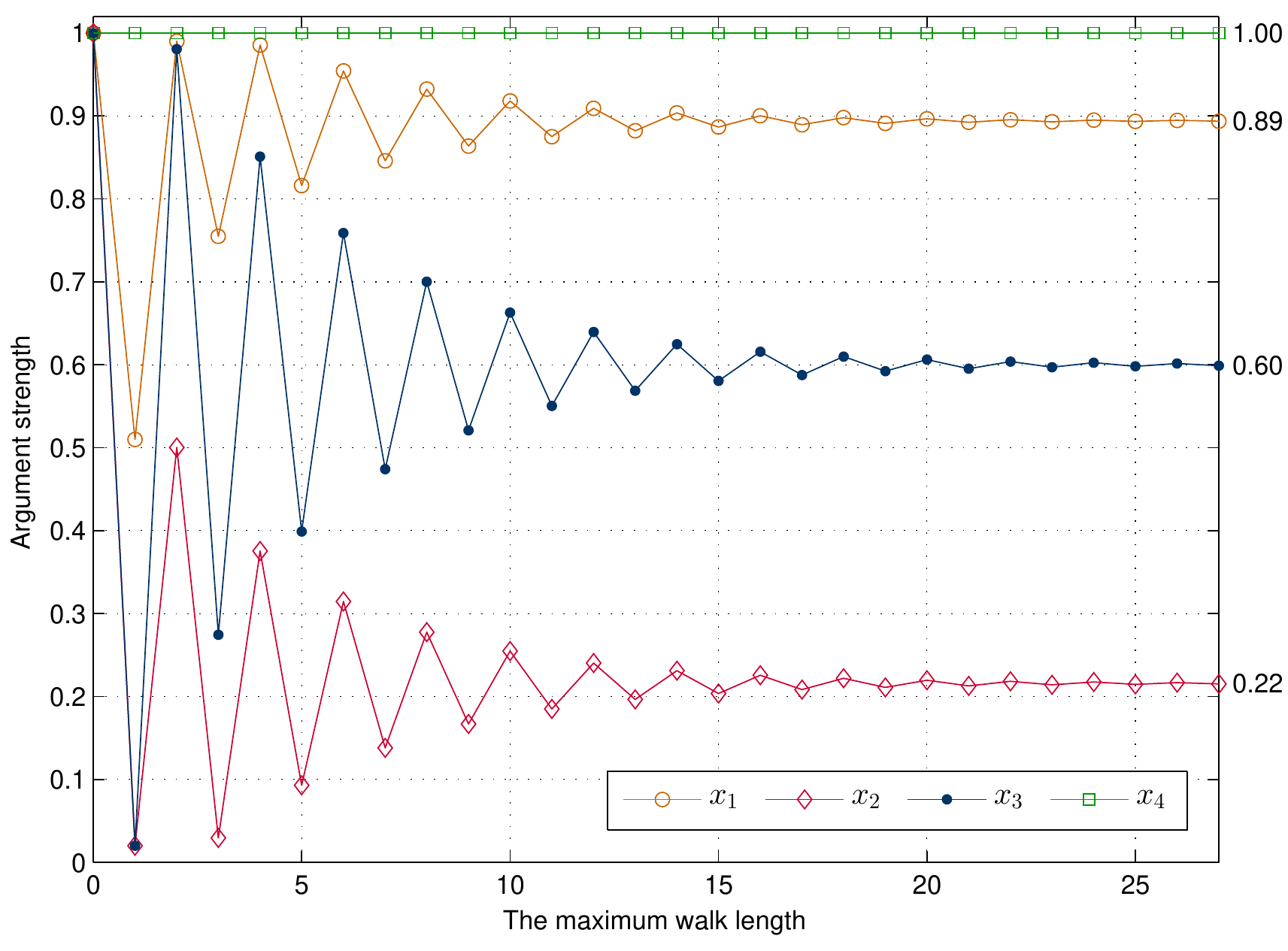}
\caption{Valuation sequence of the attacker and defender counting model for Example~\ref{Exp_SimpleAF}.}
\label{Fig_IterValuation}
\vspace{-4mm}
\end{figure}
\end{example}

\section{Some Properties of Counting Semantics}
In this section, we will give some general properties about the attacker and defender counting semantics.

\subsection{Abstraction}
The first fundamental property is called \emph{abstraction}, which corresponds to the fact that the counting semantics only depends on the attack relation between arguments while it is thoroughly independent of any characteristic of arguments at the underlying language level \cite{ref-amgoud2013ranking}. Formally, this property corresponds to the fact that argumentation frameworks which are isomorphic have the ``same'' (modulo the isomorphism) counting semantics, as stated by the following definitions:
\begin{definition}[Isomorphism]
  Two argumentation frameworks $\textit{AF}_1=\left< \mathcal{X}_1, \mathcal{R}_1\right>$ and $\textit{AF}_2=\left< \mathcal{X}_2, \mathcal{R}_2\right>$ are \emph{isomorphic} iff there exists a bijective function $\tau$: $\mathcal{X}_1 \mapsto \mathcal{X}_2$ such that for any $x,y\in \mathcal{X}_1$, $x\mathcal{R}_1 y$ iff $\tau(x)\mathcal{R}_2\, \tau(y)$.
\end{definition}
\begin{theorem}  \label{Thrm_Isomorphism}
  Let $\bm{v}^{\textit{AF}_1}_\alpha$ and $\bm{v}^{\textit{AF}_2}_\alpha$ be the attacker and defender counting semantics of $\textit{AF}_1$ and $\textit{AF}_2$ respectively, under a given damping factor $\alpha$. For any isomorphism $\tau$ from $\textit{AF}_1$ to $\textit{AF}_2$ and for any $x\in\mathcal{X}_1$, it holds that $\bm{v}^{\textit{AF}_1}_\alpha(x)=\bm{v}^{\textit{AF}_2}_\alpha\left(\tau(x)\right)$.
\end{theorem}
With this result, we have the following corollary about the argumentation framework whose argument graph is an elementary cycle:
\begin{corollary}
If the argument graph $\mathbb{G}$ of an $\textit{AF}=\left< \mathcal{X}, \mathcal{R}\right>$ is an elementary cycle, then for any arguments $x,y\in\mathcal{X}$, we have $\bm{v}_\alpha(x)=\bm{v}_\alpha(y)$.
\end{corollary}

We have stated that the strength values of arguments are relative and do not make sense when they are not compared with each other. Actually, in most applications, we merely concern the ranking (ordering) over arguments induced by the counting semantics. Given the damping factor $\alpha$, the \emph{ranking} $\succeq_\alpha$ on the set of arguments $\mathcal{X}$ derived from the counting semantics $\bm{v}_\alpha$ is defined by: for any $x,y\in\mathcal{X}$, $x\succeq_\alpha y$ iff $\bm{v}_\alpha(x)\geq\bm{v}_\alpha(y)$. Intuitively, $\succeq_\alpha$ is total (i.e., $\forall x,y\in\mathcal{X}$, $x\succeq_\alpha y$ or $y\succeq_\alpha x$) and transitive (i.e., $\forall x,y,z\in\mathcal{X}$, if $x\succeq_\alpha y$ and $y\succeq_\alpha z$, then $x\succeq_\alpha z$). Note that here $x\succeq_\alpha y$ means that argument $x$ is at least as acceptable as argument $y$ w.r.t. $\alpha$. Formally, we define $x \simeq_\alpha y$ if and only if $x\succeq_\alpha y$ and $y\succeq_\alpha x$, which means $x$ and $y$ are equally acceptable w.r.t. $\alpha$. Moreover, $x \succ_\alpha y$, meaning $x$ is strictly more acceptable than $y$ w.r.t. $\alpha$, if and only if $x \succeq_\alpha y$ but not $y \succeq_\alpha x$.
\begin{corollary}
  Assume $\textit{AF}_1=\left< \mathcal{X}_1, \mathcal{R}_1\right>$ and $\textit{AF}_2=\left< \mathcal{X}_2, \mathcal{R}_2\right>$ be isomorphic w.r.t. $\tau$, for a given damping factor $\alpha$, then we have $\forall x,y\in\mathcal{X}_1$, $x\succeq_\alpha^{\textit{AF}_1} y$ iff $\tau(x)\succeq_\alpha^{\textit{AF}_2} \tau(y)$.
\end{corollary}

Actually, this corollary is equivalent to Theorem~\ref{Thrm_Isomorphism}, and states that two isomorphic argument graphs give rise to two equivalent rankings on  arguments.

\subsection{Damping-independent ranking}
Different damping factor $\alpha$ may affect the results of the counting semantics, and thus may give the different ranking on arguments. More specifically, for an argumentation framework $\left< \mathcal{X}, \mathcal{R}\right>$, for two different damping factors $\alpha$ and $\alpha'$, and for two arguments $x,y\in\mathcal{X}$, the counting semantics $\bm{v}_\alpha$ may give that $\bm{v}_\alpha(x)\geq\bm{v}_\alpha(y)$, i.e. $x\succeq_\alpha y$, while the semantics $\bm{v}_{\alpha'}$ may give the opposite result $\bm{v}_{\alpha'}(y)\geq\bm{v}_{\alpha'}(x)$, i.e. $y\succeq_{\alpha'} x$. To investigate how different $\alpha$ influence the ranking on arguments is a quite complex thing, and we will discuss it in our future works. In this paper, we mainly concern on the properties which always hold for any damping factor $\alpha$.
\begin{proposition}  \label{Prop_Properties}
  Let $x_i,x_j\in\mathcal{X}$. For any damping factor $\alpha\in(0,1)$, the ranking $\succeq_\alpha$ induced by the counting semantics $\bm{v}_\alpha$ satisfies:
  \begin{enumerate}
  \item If $\mathcal{R}^-(x_i)=\emptyset$ and $\mathcal{R}^-(x_j)\neq\emptyset$, then $x_i \succ_\alpha x_j$.\label{PropP_1}
  \item If $\mathcal{R}^-(x_i)=\mathcal{R}^-(x_j)$, then $x_i \simeq_\alpha x_j$,
  \item If $\mathcal{R}^-(x_i) \subset \mathcal{R}^-(x_j)$, then $x_i \succ_\alpha x_j$.
\end{enumerate}
\end{proposition}

Property {\rm [P1]} states that non-attacked arguments are always the most acceptable and attacked arguments always have non-maximal valuation. This property is common in many proposals \cite{ref-cayrol2005graduality,ref-matt2008game,ref-pfa2014caterank}. Property {\rm [P2]} shows that two arguments with the same $1$-length attackers always have the same valuation (and thus are always equally acceptable). Property {\rm [P3]} reveals that an argument $x_i$, whose $1$-length attackers pertain to the set of $1$-length attackers of argument $x_j$, is always more acceptable than $x_j$. Using these properties, we can easily identify some rankings between arguments regardless of the damping factor $\alpha$.
%

\begin{example}  \label{Exp_Properties}
  Consider again the argument graph shown in Figure~\ref{Fig_AAF} where $\mathcal{R}^-(x_1)=\{x_2\}$, $\mathcal{R}^-(x_2)=\{x_3,x_4\}$, $\mathcal{R}^-(x_3)=\{x_2,x_3\}$ and $\mathcal{R}^-(x_4)=\emptyset$. Obviously, $x_4$ has the highest rank; $x_1\succ x_3$ since $\mathcal{R}^-(x_1)\subset\mathcal{R}^-(x_3)$. Then, we have the rankings: $x_4\succ x_1 \succ x_3$ and $x_4 \succ x_2$.
\end{example}

One strong result generalizes Proposition~\ref{Prop_Properties} in two ways: first it considers arbitrary number of $1$-length attackers and second, it considers various strengths of arguments. This involves a relation that compares sets of arguments, i.e. \emph{set comparison}: Let $\sqsubseteq_\alpha$ be a ranking on set $\mathcal{X}$ of arguments with respect to $\alpha$ and let $S_1,S_2\subseteq\mathcal{X}$, $S_1\sqsubseteq_\alpha S_2$ iff there is an injective mapping $\lambda$ from $S_1$ to $S_2$ such that for all $x\in S_1$, $\lambda(x)\succeq_\alpha x$. Obviously, if $S_1\sqsubseteq_\alpha S_2$, there must be $|S_1|\leq|S_2|$ and for any $x\in S_1$, there exists an argument $y$ in $S_2$ such that $y\succeq_\alpha x$.

\begin{theorem} \label{Thrm_WeakTransitive}
Let $\bm{v}_\alpha$ be an attacker and defender counting semantics w.r.t the damping factor $\alpha$. For any $x_i,x_j\in\mathcal{X}$, if $\mathcal{R}^-(x_i) \sqsubseteq_\alpha \mathcal{R}^-(x_j)$, then it holds that $x_i \succeq_\alpha x_j$.
\end{theorem}
This theorem tells us that argument $x_i$ is at least as acceptable as argument $x_j$, when the $1$-length attackers of $x_j$ at least as numerous and well-ranked as those of $x_i$. The relation of set comparison between $S_1$ and $S_2$ is \emph{strong}, denoted by $S_1\sqsubset_\alpha S_2$, iff it satisfies two conditions: (1) $S_1\sqsubseteq_\alpha S_2$; (2) $|S_1|<|S_2|$ or for some $x\in S_1$ such that $\lambda(x)\succ_\alpha x$ and $\lambda(x) \not\simeq_\alpha x$. Then, we have the strong version of Theorem~\ref{Thrm_WeakTransitive}:
\begin{theorem} \label{Thrm_StrongTransitive}
Let $\bm{v}_\alpha$ be a counting semantics w.r.t the damping factor $\alpha$. For any $x_i,x_j\in\mathcal{X}$, if $\mathcal{R}^-(x_i) \sqsubset_\alpha \mathcal{R}^-(x_j)$, then it holds that $x_i \succ_\alpha x_j$.
\end{theorem}
\begin{example}[continues=Exp_Properties]
  Now, let us compare arguments $x_2$ and $x_3$. Intuitively, $|\mathcal{R}^-(x_2)|=|\mathcal{R}^-(x_3)|$. We define the injective mapping $\lambda$ from $\mathcal{R}^-(x_3)$ to $\mathcal{R}^-(x_2)$ as: $\lambda(x_3|_{\mathcal{R}^-(x_3)})=x_3|_{\mathcal{R}^-(x_2)}$ and $\lambda(x_2|_{\mathcal{R}^-(x_3)})=x_4|_{\mathcal{R}^-(x_2)}$, where $x|_S$ means the element $x$ in set $S$. 
  Based on the previous rankings: $x_3|_{\mathcal{R}^-(x_2)} \simeq x_3|_{\mathcal{R}^-(x_3)}$ and $x_4|_{\mathcal{R}^-(x_2)}\succ x_2|_{\mathcal{R}^-(x_3)}$, we have $\mathcal{R}^-(x_3)\sqsubset \mathcal{R}^-(x_2)$, and by Theorem~\ref{Thrm_StrongTransitive}, we have $x_3 \succ x_2$. Then, we can conclude the ranking on all arguments in Figure~\ref{Fig_AAF}: $x_4\succ x_1 \succ x_3 \succ x_2$, which are consistent with the results in Example~\ref{Exp_ValuationSeq}.
%
%
%
\end{example}

\section{Related Work and Conclusion}
This paper mainly focuses on evaluating arguments by assigning a strength to each argument. In this regard, there exists numerous works \cite{ref-cayrol2005graduality,ref-matt2008game,ref-leite2011SocialAF,ref-gabbay2012equational,ref-pfa2014caterank}, etc. However the most related works may be the gradual approach in \cite{ref-cayrol2005graduality} and the equational approach in \cite{ref-gabbay2012equational} since both these two approaches and our counting approach can be seen as interaction-based approaches, i.e., evaluating arguments based on the graph structure of the argumentation framework. Our model can be seen as a linear model and has significant computational advantages.

In the short term, future work mainly aim to the following aspects. First, the damping factor plays an important role in our counting semantics. How the damping factor influences the results and how to decide it are two urgent problems. Second, argumentation has become social activities by Web 2.0 in our daily life \footnote{The websites \url{www.livingvote.org}, \url{debategraph.org}, \url{idebate.org} are a few examples.}. How to extend our work to evaluate arguments in social context is another research point.


{  
\small
\bibliographystyle{IEEEtran}

\bibliography{sigproc}  

\begin{thebibliography}{10}
\providecommand{\url}[1]{#1}
\csname url@samestyle\endcsname
\providecommand{\newblock}{\relax}
\providecommand{\bibinfo}[2]{#2}
\providecommand{\BIBentrySTDinterwordspacing}{\spaceskip=0pt\relax}
\providecommand{\BIBentryALTinterwordstretchfactor}{4}
\providecommand{\BIBentryALTinterwordspacing}{\spaceskip=\fontdimen2\font plus
\BIBentryALTinterwordstretchfactor\fontdimen3\font minus
  \fontdimen4\font\relax}
\providecommand{\BIBforeignlanguage}[2]{{%
\expandafter\ifx\csname l@#1\endcsname\relax
\typeout{** WARNING: IEEEtran.bst: No hyphenation pattern has been}%
\typeout{** loaded for the language `#1'. Using the pattern for}%
\typeout{** the default language instead.}%
\else
\language=\csname l@#1\endcsname
\fi
#2}}
\providecommand{\BIBdecl}{\relax}
\BIBdecl

\bibitem{ref-rahwan2009argumentation}
I.~Rahwan and G.~R. Simari, \emph{Argumentation in artificial
  intelligence}.\hskip 1em plus 0.5em minus 0.4em\relax Springer, 2009.

\bibitem{ref-Dung1995AAF}
P.~M. Dung, ``On the acceptability of arguments and its fundamental role in
  nonmonotonic reasoning, logic programming and n-person games,'' \emph{Journal
  of Artificial Intelligence}, vol.~77, no.~2, pp. 321--357, Sep. 1995.

\bibitem{ref-semantic:KER}
P.~Baroni, M.~Caminada, and M.~Giacomin, ``An introduction to argumentation
  semantics,'' \emph{The Knowledge Engineering Review}, vol.~26, pp. 365--410,
  12 2011.

\bibitem{ref-bench2007ArgAI}
T.~J. Bench-Capon and P.~E. Dunne, ``Argumentation in artificial
  intelligence,'' \emph{Artificial intelligence}, vol. 171, no.~10, pp.
  619--641, 2007.

\bibitem{ref-cayrol2005graduality}
C.~Cayrol and M.-C. Lagasquie-Schiex, ``Graduality in argumentation,''
  \emph{Journal Artificial Intelligence Research (JAIR)}, vol.~23, pp.
  245--297, 2005.

\bibitem{ref-matt2008game}
P.-A. Matt and F.~Toni, ``A game-theoretic measure of argument strength for
  abstract argumentation,'' in \emph{Logics in Artificial Intelligence}.\hskip
  1em plus 0.5em minus 0.4em\relax Springer, 2008, pp. 285--297.

\bibitem{ref-leite2011SocialAF}
J.~Leite and J.~Martins, ``Social abstract argumentation,'' in
  \emph{Proceedings of the Twenty-Second international joint conference on
  Artificial Intelligence-Volume Volume Three}.\hskip 1em plus 0.5em minus
  0.4em\relax AAAI Press, 2011, pp. 2287--2292.

\bibitem{ref-gabbay2012equational}
D.~M. Gabbay, ``Equational approach to argumentation networks,'' \emph{Argument
  \& Computation}, vol.~3, no. 2-3, pp. 87--142, 2012.

\bibitem{modgil2013added}
S.~Modgil, F.~Toni, F.~Bex, I.~Bratko, C.~I. Ches{\~n}evar,
  W.~Dvo{\v{r}}{\'a}k, M.~A. Falappa, X.~Fan, S.~A. Gaggl, A.~J. Garc{\'\i}a
  \emph{et~al.}, ``The added value of argumentation,'' in \emph{Agreement
  Technologies}.\hskip 1em plus 0.5em minus 0.4em\relax Springer, 2013, pp.
  357--403.

\bibitem{ref-amgoud2013ranking}
L.~Amgoud and J.~Ben-Naim, ``Ranking-based semantics for argumentation
  frameworks,'' in \emph{Scalable Uncertainty Management}.\hskip 1em plus 0.5em
  minus 0.4em\relax Springer, 2013, pp. 134--147.

\bibitem{ref-pfa2014caterank}
F.~Pu, J.~Luo, Y.~Zhang, and G.~Luo, ``Argument ranking with categoriser
  function,'' in \emph{Knowledge Science, Engineering and Management}.\hskip
  1em plus 0.5em minus 0.4em\relax Springer, 2014, pp. 290--301.

\bibitem{Simari2009argame}
S.~Modgil and M.~Caminada, ``\BIBforeignlanguage{English}{Proof theories and
  algorithms for abstract argumentation frameworks},'' in
  \emph{\BIBforeignlanguage{English}{Argumentation in Artificial
  Intelligence}}, G.~Simari and I.~Rahwan, Eds.\hskip 1em plus 0.5em minus
  0.4em\relax Springer US, 2009, pp. 105--129.

\bibitem{ref-rienstra2013opponent}
T.~Rienstra, M.~Thimm, and N.~Oren, ``Opponent models with uncertainty for
  strategic argumentation,'' in \emph{Proceedings of the Twenty-Third
  international joint conference on Artificial Intelligence}.\hskip 1em plus
  0.5em minus 0.4em\relax AAAI Press, 2013, pp. 332--338.

\bibitem{ref-horn2012matrix}
R.~A. Horn and C.~R. Johnson, \emph{Matrix analysis}.\hskip 1em plus 0.5em
  minus 0.4em\relax Cambridge university press, 2012.

\end{thebibliography}
} 
\end{document}